\documentclass[sigconf]{acmart}
\AtBeginDocument{%
  }
  
\hypersetup{
  pdfauthor={Ximiao Li, Lin Jiang, Rongchao Xu, Dahai Yu, Zhe He, Guang Wang},
  pdftitle={SynEHR: Joint Modeling Inter-visit Temporal Evolution and Intra-visit Clinical Structure for Longitudinal EHR Synthesis},
  pdfsubject={},
  pdfkeywords={}
}

\copyrightyear{2026}
\acmYear{2026}
\setcopyright{cc}
\setcctype{by}
\acmConference[CIKM '26]{Proceedings of the 35th ACM International
Conference on Information and Knowledge Management}{November 07--11,
2026}{Rome, Italy}
\acmBooktitle{Proceedings of the 35th ACM International Conference on
Information and Knowledge Management (CIKM '26), November 07--11, 2026,
Rome, Italy}
\acmDOI{10.1145/3799682.3840734}
\acmISBN{979-8-4007-2539-5/2026/11}
\usepackage[T1]{fontenc}
\usepackage{amsfonts}
\usepackage{amsmath}
\usepackage{amsthm}
\usepackage{graphicx}
\usepackage{subcaption}
\usepackage{epstopdf}
\usepackage{enumitem}
\usepackage{multirow}
\usepackage{algorithmic}
\usepackage{amsopn}
\usepackage{xspace}
\usepackage{xcolor}
\usepackage{hyperref}

\setlist[itemize]{topsep=3pt, itemsep=1pt, parsep=0pt, partopsep=0pt}
\setlist[enumerate]{topsep=3pt, itemsep=1pt, parsep=0pt, partopsep=0pt}

\ifpdf
  \DeclareGraphicsExtensions{.pdf,.png,.jpg,.eps}
\else
  \DeclareGraphicsExtensions{.eps}
\fi

\theoremstyle{remark}

\theoremstyle{plain}

\begin{document}

\newcommand{\N}{SynEHR\xspace}

\title{\N: Joint Modeling Inter-visit Temporal Evolution and Intra-visit Clinical Structure for Longitudinal EHR Synthesis}

\author{Ximiao Li}
% \authornote{Corresponding authors.}
\affiliation{%
  \institution{Florida State University}
  \city{Tallahassee}
  \state{Florida}
  \country{United States}
}
\email{xl24g@fsu.edu}

\author{Lin Jiang}
\affiliation{%
  \institution{Florida State University}
  \city{Tallahassee}
  \state{Florida}
  \country{United States}
}
\email{lj23d@fsu.edu}

\author{Rongchao Xu}
\affiliation{%
  \institution{Florida State University}
  \city{Tallahassee}
  \state{Florida}
  \country{United States}
}
\email{rx21a@fsu.edu}

\author{Dahai Yu}
\affiliation{%
  \institution{Florida State University}
  \city{Tallahassee}
  \state{Florida}
  \country{United States}
}
\email{dahai.yu@fsu.edu}

\author{Zhe He}
\affiliation{%
  \institution{Florida State University}
  \city{Tallahassee}
  \state{Florida}
  \country{United States}
}
\email{zhe@fsu.edu}

\author{Guang Wang}
\authornote{Corresponding author.}
\affiliation{%
  \institution{Florida State University}
  \city{Tallahassee}
  \state{Florida}
  \country{United States}
}
\email{guang@cs.fsu.edu}

\renewcommand{\shortauthors}{Li et al.}

\begin{abstract}
Longitudinal electronic health records (EHRs) document patients' sequences of clinical visits over time, preserving the temporal evolution of disease progression and care delivery. However, real longitudinal EHRs are difficult to access because they contain large amounts of fine-grained, patient-specific information. Synthetic EHR generation therefore provides a valuable approach for preserving the statistical patterns and clinical structure of patient visit trajectories, enabling broader modeling and analysis when real records are limited. Although recent generative models have made progress in producing future visit sequences, they remain limited in explicitly integrating inter-visit irregular temporal evolution and intra-visit clinical event structures in EHRs, leading to clinically inconsistent and temporally unrealistic visit sequences. In this work, we propose~\N, a lightweight adaptive LLM-based framework for longitudinal EHR synthesis. There are two novel designs in \N, i.e., a Temporal State Conditioning Module captures irregular temporal states across visits and a Temporal-Relational Adaptation Module combines these states with patient history to dynamically construct patient-specific relational representations. \N then builds on a parameter-efficient LoRA-adapted language-model generator with next-visit generation capability to train the two modules for temporally and clinically informed generation. Extensive experiments on real-world EHR datasets across fidelity, privacy, and downstream utility evaluations demonstrate that \N outperforms state-of-the-art models by generating more clinically coherent and temporally faithful longitudinal EHR data.

\end{abstract}




\maketitle

\section{Introduction}

Longitudinal electronic health records (EHRs) have become widely adopted in modern clinical care, providing a temporal view of patients’ disease progression and care delivery~\cite{onc2022ehr}. These trajectories consist of successive visits associated with visit timestamps and clinical content from multiple data types, including diagnoses, procedures, medications, laboratory findings, and other structured observations. However, the granular and temporally linked nature of longitudinal EHR data introduces substantial privacy, governance, and re-identification challenges, limiting the broad sharing of fine-grained patient-level records. Synthetic longitudinal EHR generation therefore offers a promising approach for supporting downstream clinical research and modeling, including proactive care planning, early disease-progression assessment, and large-scale clinical analysis~\cite{choi2016retain,rajkomar2018scalable} while reducing reliance on direct sharing of sensitive patient data.
To be clinically meaningful, such synthetic EHR data should preserve both temporal evolution across successive visits and structured relations among clinical events within each visit~\cite{shickel2018deep}. For example, an EHR trajectory may contain closely related diagnosis and medication records for the same disease within a visit, while the content of a subsequent visit often remains closely connected to the preceding clinical history.

Existing studies on longitudinal EHR synthesis can be broadly categorized into three directions: (i) predictive EHR models learn patient representations from historical records, but they are designed for task-specific clinical prediction rather than longitudinal visit trajectory synthesis~\cite{choi2016retain,li2020behrt,rasmy2021medbert,chen2024trans};(ii) generative EHR models can autoregressively produce future visit sequences from prior histories, but existing textual or event-token sequential formulations may not explicitly preserve both clinical context and temporally coherent transitions across multiple visits~\cite{renc2024ethos,makarov2025dtgpt}; and (iii) LLM-based approaches have recently advanced clinical and EHR task modeling, yet they largely focus on task-level applications and do not explicitly support dynamic structural modeling of clinical context for patient-specific longitudinal EHR synthesis~\cite{li2024llamacare,gema2024peftllama,hasheminasab2026ehrfinetune}. Consequently, existing approaches still lack a flexible and explicit mechanism for integrating \textbf{inter-visit temporal evolution} with \textbf{intra-visit clinical structure} in longitudinal EHR.

However, jointly modeling visit-level temporal evolution and event relations remains challenging. First, inter-visit temporal evolution is difficult to capture reliably, as longitudinal EHR trajectories contain highly irregular visit intervals that can substantially influence subsequent states and clinical content. Simply using these intervals to guide generation may therefore provide unstable or insufficient temporal information. Second, intra-visit clinical structure is challenging to model explicitly because it is patient-specific and context-dependent rather than fixed. The relations among different data domains within a visit may vary with the patient’s prior clinical history and evolving temporal state. As a result, static or globally shared event patterns are insufficient to guide clinically coherent trajectory generation.

To address these challenges, we propose~\N, a lightweight adaptive LLM-based framework for longitudinal EHR synthesis by joint modeling inter-visit temporal evolution and
intra-visit clinical structure. \N consists of two novel modules, i.e., Temporal State Conditioning Module (TSCM) and Temporal-Relational Adaptation Module (TRAM) to explicitly model inter-visit temporal evolution and intra-visit clinical structure for adaptation. 
(i) TSCM is designed to capture the inter-visit temporal evolution reflected by irregular visit intervals, which includes a Temporal Feature Encoding to summarize temporal patterns and visit-level clinical context from the historical trajectory into a patient-specific latent temporal state, and a Temporal State Head to convert this representation into confidence-aware conditioning signals, providing a more stable and informative temporal context for subsequent generation and relational adaptation.
(ii) TRAM models patient-specific intra-visit clinical structure through two complementary branches. The Static Branch summarizes patient history to capture core relations across data types, while the Dynamic Branch adapts these relations using the temporal context from TSCM and the current patient state. The two branches are fused into a patient-specific relational representation and converted into soft prefix tokens to guide next-visit generation.
In addition, \N uses a parameter-efficient LoRA-adapted LLM generator to autoregressively synthesize future visits, including both visit timestamps and clinical content. For each EHR dataset, \N keeps the generator fixed and only trains TSCM and TRAM. This design enables adaptation for context-aware temporal and structural conditioning without requiring full retraining of the LLM generator.
% The contributions of~\N can be summarized as follows:

The main contributions of this paper are as follows:
\begin{itemize}[leftmargin=*]
    \item \textbf{Conceptually}, we propose~\N, a lightweight adaptive LLM-based framework for longitudinal EHR synthesis. Rather than treating future visits as unstructured sequences of clinical events, \N explicitly models both \emph{inter-visit temporal evolution} and \emph{intra-visit clinical structure}, enabling more temporally faithful and clinically coherent patient trajectory generation.
    \item \textbf{Technically}, there are two key innovative components in \N. A \textbf{TSCM} models irregular inter-visit temporal evolution and provides stable temporal conditioning, and a \textbf{TRAM} learns patient-specific intra-visit clinical relations and converts them into soft prefix guidance for the LLM-based generator. Integration of these two components allows~\N to model evolving visit patterns and clinical dependencies in a unified manner.
    \item \textbf{Experimentally}, we extensively evaluate~\N on two real-world critical-care EHR datasets, i.e., MIMIC-III and MIMIC-IV, against state-of-the-art baselines in terms of synthesis fidelity, privacy preservation, and downstream clinical utility. The results demonstrate that~\N delivers strong overall performance across these evaluation dimensions. For example, on MIMIC-III,~\N improves diagnosis, medication, and procedure longitudinal fidelity by 2.5\%, 3.0\%, and 0.9\%, respectively, while reducing the discrepancy in inter-visit time distributions by 10.9\%. The code of this paper is available at \textcolor{blue}{\textbf{\url{https://github.com/MiaL7/SynEHR}}}.
\end{itemize}

\section{Data Analysis and Motivation}
\label{sec:motivation}

In this section, we conduct a data-driven analysis to motivate two key factors in longitudinal EHR synthesis: inter-visit temporal evolution and intra-visit clinical structure. These two factors are closely coupled in longitudinal EHRs. On the one hand, the clinical contexts reflected in previous visits can influence when the next visit occurs. On the other hand, the resulting inter-visit temporal evolution provides important context for the next visit, as different inter-visit intervals often correspond to different visit-level clinical contexts.

\begin{figure}[t]
  \centering
  \includegraphics[width=0.85\linewidth]{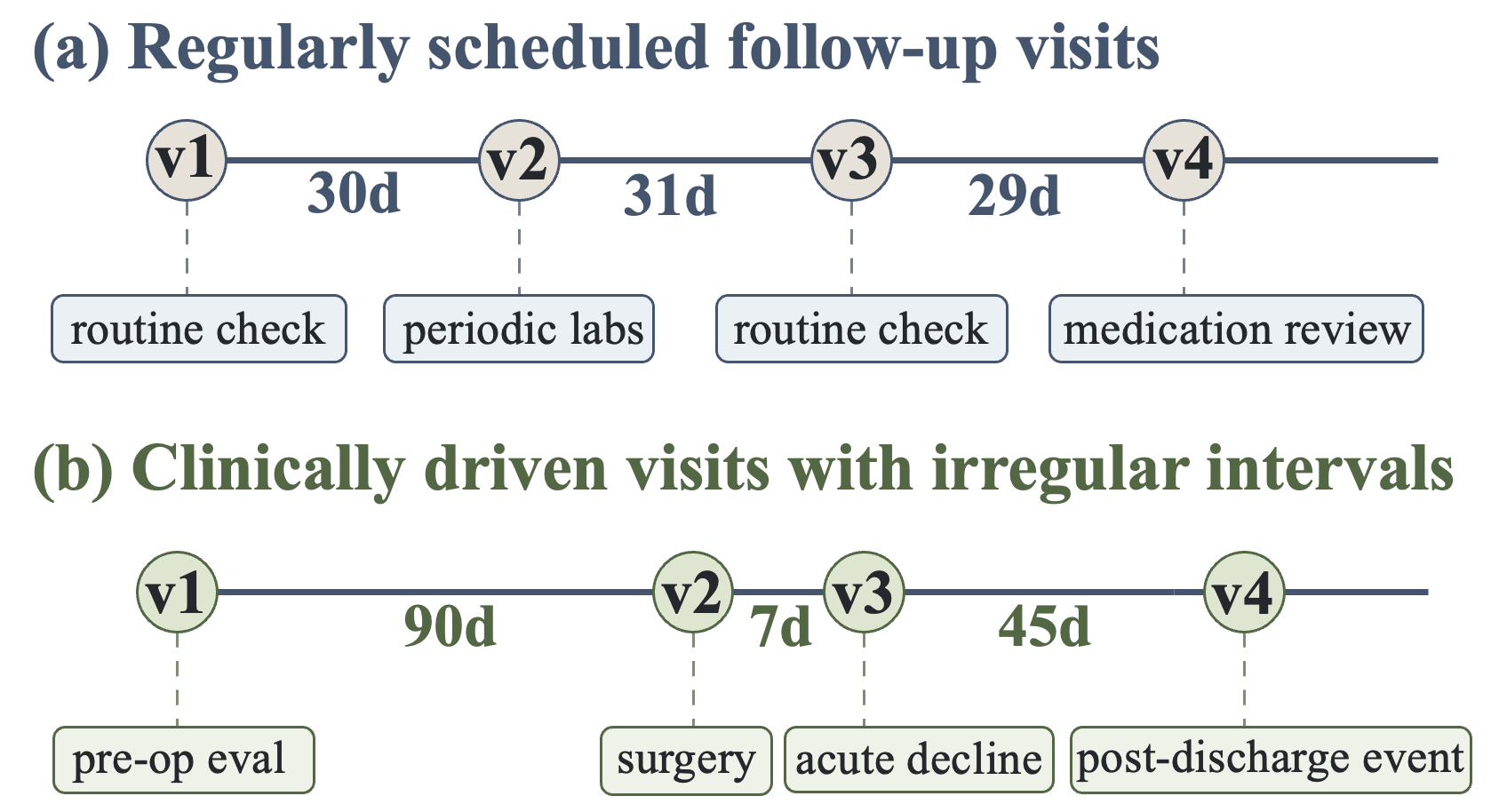}
  \caption{Illustration of regular and irregular visit patterns.}
  \label{fig:temporal_uncertainty_patient_lines}
\end{figure}

\begin{figure*}[t]
  \centering
  \begin{subfigure}[t]{0.33\textwidth}
    \centering
    \includegraphics[width=\linewidth]{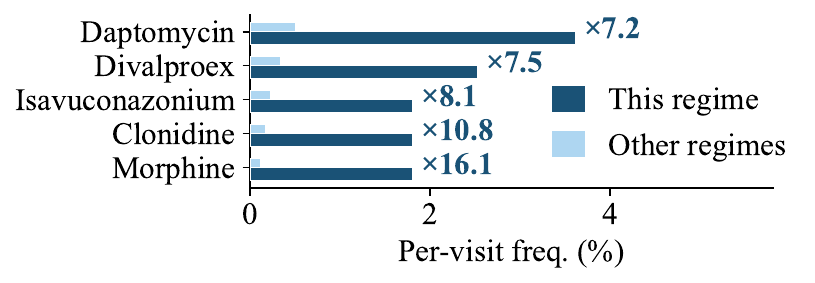}
    \caption{Short Regime (Acute Care, $n{=}277$)}
    \label{fig:tram_drift_short}
  \end{subfigure}
  \hfill
  \begin{subfigure}[t]{0.33\textwidth}
    \centering
    \includegraphics[width=\linewidth]{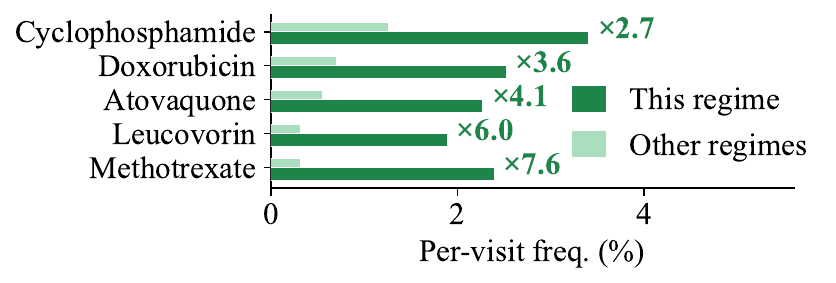}
    \caption{Medium Regime (Scheduled Care, $n{=}792$)}
    \label{fig:tram_drift_medium}
  \end{subfigure}
  \hfill
  \begin{subfigure}[t]{0.33\textwidth}
    \centering
    \includegraphics[width=\linewidth]{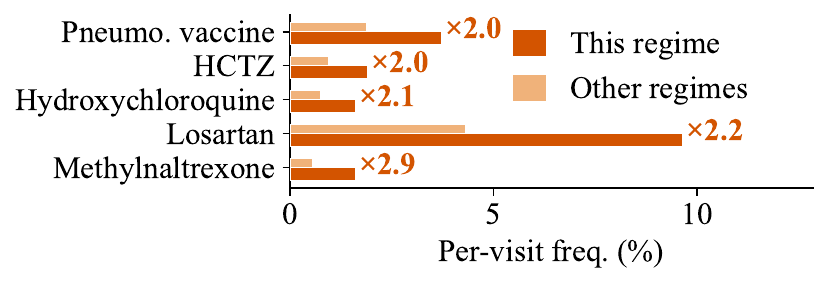}
    \caption{Long Regime (Surveillance, $n{=}995$)}
    \label{fig:tram_drift_long}
  \end{subfigure}
  \caption{%
    Temporal-regime-specific medication enrichment among visits with
    Malignant Neoplasm (CCSR NEO072) in the MIMIC-IV dataset.
  }
  \label{fig:tram_drift}
\end{figure*}

\subsection{Inter-visit Temporal Evolution}

\textbf{Inter-visit temporal evolution in longitudinal EHRs is highly irregular and clinically informative, since the timing of future visits is closely associated with the clinical contexts observed in patients' historical trajectories.} As illustrated in Figure~\ref{fig:temporal_uncertainty_patient_lines}, different patients may exhibit distinct temporal patterns under different clinical scenarios. For example, Patient (a) follows a regular follow-up schedule, with visits such as routine checks and laboratory tests occurring at stable intervals. In contrast, Patient (b) shows a more irregular pattern, where visit intervals change substantially around clinical events such as surgery, acute decline, or post-discharge events. These examples suggest that inter-visit intervals should not be treated as simple timestamp differences, but rather as temporal signals that are closely associated with visit-level clinical contexts and reflect the evolution of patient states and care processes. Accurately modeling inter-visit temporal evolution is therefore important for generating longitudinal EHR trajectories that are temporally faithful and clinically coherent.

\subsection{Intra-visit Clinical Structure}

\textbf{Intra-visit clinical structure is inherently complex and shaped by inter-visit temporal evolution, since different inter-visit intervals can lead to different visit-level clinical contexts in the next visit.} We examine this effect using Malignant Neoplasm (CCSR NEO072) as a representative anchor diagnosis. As shown in Figure~\ref{fig:tram_drift}, visits with the same diagnosis exhibit different medication patterns across short-, medium-, and long-interval temporal regimes, defined as $\leq 7$ days, 8--90 days, and $>90$ days, respectively: short-interval visits are associated with acute-care drugs, medium-interval visits with chemotherapy-related medications, and long-interval visits with surveillance and chronic-management medications. This finding suggests that intra-visit clinical structure is not determined by diagnosis alone, but also shaped by temporal evolution, highlighting the need for temporally adaptive structure modeling in longitudinal EHR synthesis.

\section{Preliminaries}
\label{sec:preliminaries}
% This section defines the notation, task setting, patient trajectory representation, and the background concepts required for the proposed model.
In this section, we formally define the task of longitudinal EHR synthesis and introduce the prefix-tuning used in our framework.

\subsection{Problem Definition}
A longitudinal EHR record of a patient is defined as a visit sequence
\begin{equation}
\mathcal{X} = [x_1, x_2, \dots, x_N],
\end{equation}
where $N$ is the total number of visits, and $x_i = (V_i, T_i)$ denotes the $i$-th visit, with $V_i$ denoting the visit-level clinical content and $T_i$ the corresponding visit timestamp. The clinical content of each visit is represented as a set of data-type-specific code sets,
\begin{equation}
V_i = \{M_i^{(1)}, M_i^{(2)}, \dots, M_i^{(K)}\},
\end{equation}
where $K$ is the number of clinical data types, such as diagnoses, procedures, medications, and laboratory items. 
Let \(\mathcal{M}=\{1,\dots,K\}\) denote the clinical data-type index set.
$M_i^{(k)} \subseteq \mathcal{C}^{(k)}$ denotes the set of observed clinical codes in the $k$-th data type for \(k\in\mathcal{M}\), where each code corresponds to a structured clinical concept from the vocabulary $\mathcal{C}^{(k)}$. Given the historical visits of a patient,
\begin{equation}
    \mathcal{X}_{1:n-1} = [x_1, x_2, \dots, x_{n-1}],
\end{equation}
the goal of EHR synthesis is to generate the next visit
$
\hat{x}_n = (\hat{V}_n, \hat{T}_n),
$
where \(\hat{V}_n\) and
\(\hat{T}_n\) denotes the generated clinical content and timestamp of the next visit, respectively. By repeating the next-visit generation process, the model can
synthesize a longitudinal EHR trajectory. Formally, this sequential generation process can be written as
\begin{equation}
p_{\theta}(\mathcal{X}) =
\prod_{n=2}^{N} p_{\theta}(x_n \mid \mathcal{X}_{1:n-1}),
\end{equation}
where \(p_{\theta}(x_n \mid \mathcal{X}_{1:n-1})\) denotes one next-visit
generation step, taking the historical visits \(\mathcal{X}_{1:n-1}\) as input
and generating the next visit \(x_n = (V_n, T_n)\) as output.

\subsection{Prefix Tuning}
Prefix tuning is a \textbf{parameter-efficient adaptation strategy}, which augments a pretrained model with a small set of learnable continuous vectors, often interpreted as virtual prompt tokens. These vectors provide additional conditioning context for generation. At the same time, they leave the discrete input sequence unchanged and avoid full-model fine-tuning.

Let \(\mathcal{X}_{1:n-1}\) denote the historical visit sequence defined in the previous subsection, and let
\(s_{1:n-1} = \mathrm{Tokenized}(\mathcal{X}_{1:n-1})\) be its tokenized sequence. In standard prefix tuning, a soft prefix is defined as a set of \(m\) learnable embeddings
\begin{equation}
    Z = [z_1, z_2, \dots, z_m] \in \mathbb{R}^{m \times d},
\end{equation}
where \(d\) is the hidden dimension of the generator. In the standard setting, the same prefix is shared across samples within a task. It is learned to steer the pretrained model toward task-relevant behavior. Because only the prefix parameters need to be optimized, prefix tuning offers a lightweight alternative to full fine-tuning. This property makes it especially appealing when training data or computational resources are limited.

However, a shared task-level prefix is not ideal for longitudinal EHR modeling, where conditioning should vary across patient histories. Our method therefore extends standard prefix tuning with context-aware conditioning. In particular,~\N learns a shared module that maps the observed trajectory to a context-aware soft prefix. The concrete prefix injection mechanism is described later in Section~\ref{sec:generator}.

\section{Methodology}
\label{sec:method}

\begin{figure*}[t]
  \centering
  \includegraphics[width=\linewidth]{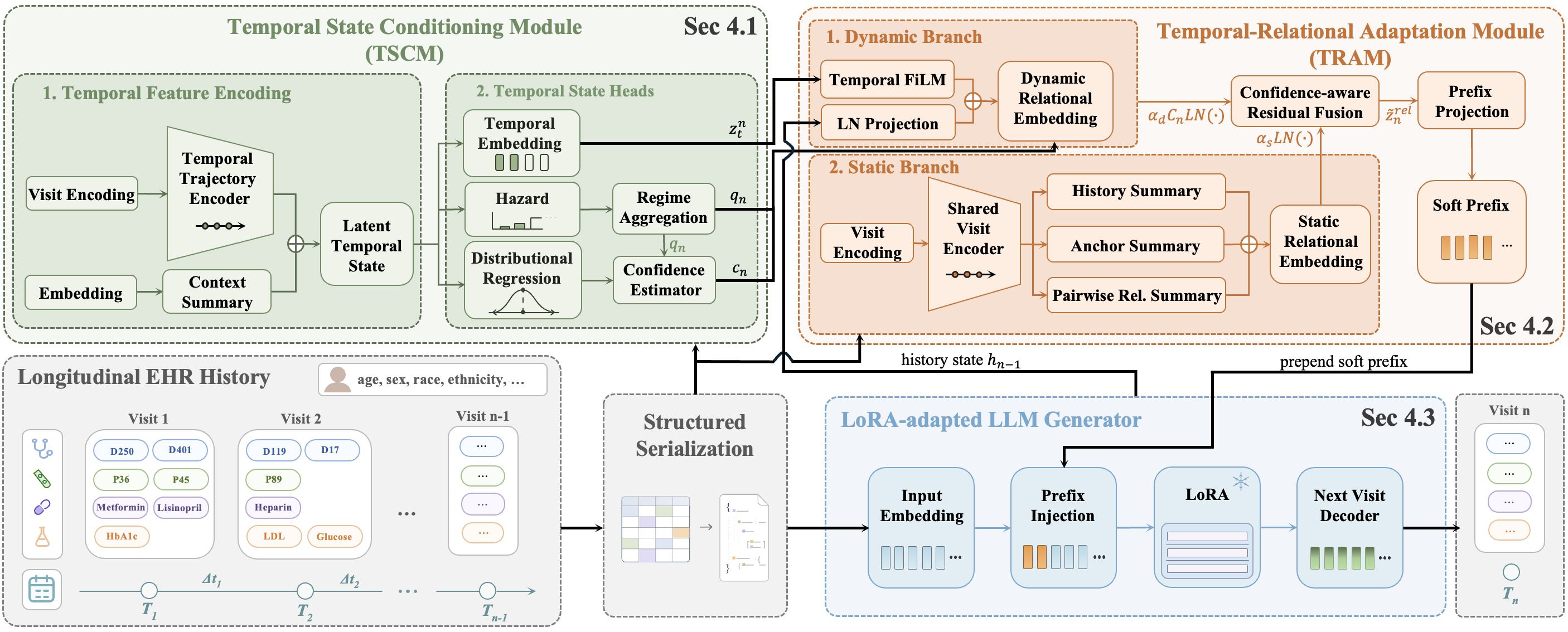}
  \caption{Overview of the~\N framework. 
    The Temporal State Conditioning Module (TSCM) encodes inter-visit intervals into a compact temporal state (Sec. 4.1), and the Temporal-Relational Adaptation Module (TRAM) captures stable and dynamic clinical relations across data types from the patient history (Sec. 4.2). The fused representation is projected into patient-specific soft prefix tokens and injected into the LoRA-adapted backbone to guide autoregressive next-visit generation (Sec. 4.3).
    }
  \label{fig:framework}
\end{figure*}

In this work, we propose~\N for longitudinal EHR synthesis. The overall framework of~\N is shown in Figure~\ref{fig:framework}. We first serialize each patient's longitudinal history into a structured token sequence, which~\N then processes with three main components, i.e., Temporal State Conditioning Module (TSCM), Temporal-Relational
Adaptation Module (TRAM), and LoRA-adapted LLM generator. 
Section~\ref{sec:TSCM} presents the TSCM, which models inter-visit temporal evolution using the patient's historical state and timing information. In Section~\ref{sec:TRAM}, we will present the TRAM, which captures intra-visit clinical structure and produces patient-specific soft prefix tokens. These tokens are injected into the generator to guide personalized EHR generation. Section~\ref{sec:generator} introduces the LoRA-adapted LLM generator, which is first fine-tuned for next-visit generation and then kept fixed during subsequent adaptation.
% while other modules generate patient-adaptive prefixes.
Finally, Section~\ref{sec:training_inference} describes the staged training procedure and autoregressive inference strategy of~\N.

\subsection{Temporal State Conditioning Module}
\label{sec:TSCM}
The Temporal State Conditioning Module (TSCM) is designed to capture inter-visit temporal evolution by converting irregular historical visit intervals into a patient-specific temporal state for next-visit generation.  As shown in Figure~\ref{fig:framework}, TSCM is organized into two stages. First, \textit{temporal feature encoding} transforms the observed visit history and inter-visit gaps into a latent temporal state that reflects the patient's evolving trajectory. Second, \textit{temporal state heads} map this latent state into three outputs: a compact temporal embedding \(z_n^t\), a coarse temporal-regime distribution \(q_n\), and a confidence score \(c_n\). Together, these signals allow SynEHR to explicitly model irregular inter-visit dynamics and provide uncertainty-aware temporal guidance for the subsequent TRAM module.

\textbf{Temporal Feature Encoding.}
For each generation step \(n\), TSCM takes the observed history \(\mathcal{X}_{1:n-1}=[x_1,\dots,x_{n-1}]\) as input and first constructs temporal features from the historical visits. 
We use the clinical data-type index set \(\mathcal{M}\) defined in Section~\ref{sec:preliminaries}, and each \(k\in\mathcal{M}\) indexes one code data type.
For visit \(i\), let \(\mathcal{C}_i^{(k)}\) denote the set of observed clinical codes in the \(k\)-th code data type, and let \(\Delta_i = T_i - T_{i-1}\) denote the inter-visit interval.
% Let \(\mathcal{M}\) denote the set of clinical code data types that constitute the visit-level clinical content, such as diagnosis, procedure, medication, and laboratory codes. For visit \(i\), let \(\mathcal{C}_i^{(k)}\) denote the set of observed clinical codes in the \(k\)-th code data type, where \(k \in \mathcal{M}\), and let \(\Delta_i = T_i - T_{i-1}\) denote the inter-visit interval. 
We summarize the clinical content of each visit by type-wise pooling, where \(e_i^{(k)}=\mathrm{MeanPool}(E^{(k)}(\mathcal{C}_i^{(k)}))\). We also encode the visit type and interval feature as \(e_i^{\mathrm{type}}=E^{\mathrm{type}}(\tau_i)\) and \(g_i=\phi_{\mathrm{gap}}([\log(1+\Delta_i),m_i])\), where \(\tau_i\) is the visit type and \(m_i \in \{0,1\}\) indicates whether the interval is missing. These features are combined into a visit-level temporal representation:
\begin{equation}
e_i^{\mathrm{code}}
= \operatorname*{Concat}_{k\in\mathcal{M}} e_i^{(k)},\qquad
v_i = \phi_{\mathrm{visit}}([e_i^{\mathrm{code}}; e_i^{\mathrm{type}}; g_i]).
\end{equation}
The sequence \(\{v_i\}_{i=1}^{n-1}\) is then fed into the \textit{Temporal Trajectory Encoder}:
\begin{equation}
r_{n-1} = \mathrm{GRU}(v_1,\dots,v_{n-1}).
\end{equation}
In parallel, we construct an auxiliary context summary \(p_{n-1}\) from the demographic embedding \(\phi_{\mathrm{demo}}(u)\), trajectory length, and most recent inter-visit gap. Finally, we fuse this auxiliary context with the sequence summary to form the \textit{Latent Temporal State}:

\begin{equation}
\ell_{n-1} = \phi_{\mathrm{state}}([r_{n-1}; p_{n-1}]).
\end{equation}
This latent state summarizes both the temporal evolution of the visit sequence and the current coarse temporal context.

\textbf{Temporal State Heads.}
Given the latent temporal state \(\ell_{n-1}\), TSCM applies three prediction heads to produce complementary temporal outputs: a temporal embedding head, a hazard head, and a distributional regression head.

The first \emph{temporal embedding head} produces a compact temporal embedding
\begin{equation}
z_n^{t} = \phi_{t}(\ell_{n-1}),
\end{equation}
which is later used by the dynamic branch of TRAM for temporal modulation.

The second \emph{hazard head} predicts a fine-grained discrete-time hazard distribution for the next interval \(\Delta_n\),
\begin{equation}
\begin{aligned}
q_n &=
[q_n^{\mathrm{short}},
q_n^{\mathrm{medium}},
q_n^{\mathrm{long}}],\\
q_n^{r}
&=
\sum_{b\in\mathcal{B}_r}
q_n^{\mathrm{fine}}(b).
\end{aligned}
\end{equation}
Here, \(r\) denotes one of the short-, medium-, and long-gap regimes, and \(\{\mathcal{B}_r\}\) partitions the \(B\) fine-grained interval bins accordingly. The fine-grained interval distribution \(q_n^{\mathrm{fine}}\) is derived from a discrete-time hazard model over \(B\) ordered temporal bins:
\begin{equation}
\begin{aligned}
q_n^{\mathrm{fine}}(b)
&=
\frac{\exp(\log p_{n,b})}
{\sum_{j=1}^{B}\exp(\log p_{n,j})},\\
\log p_{n,b}
&=
\log \lambda_{n,b}
+
\sum_{j<b}\log(1-\lambda_{n,j}),\\
\lambda_{n,b}
&=
\sigma(a_{n,b}),\\
a_n
&=
W_h \ell_{n-1} + b_h .
\end{aligned}
\end{equation}
Here, \(\lambda_{n,b}\) denotes the conditional hazard that the next visit falls into temporal interval bin \(b\), and \(q_n^{\mathrm{fine}}\) is normalized to obtain the predictive distribution over fine-grained inter-visit intervals.

The third \emph{distributional regression head} predicts the continuous interval statistics in log-day space and derives a scalar confidence score:
\begin{equation}
c_n = \frac{1}{2}c_n^{\mathrm{ent}} + \frac{1}{2}c_n^{\mathrm{scale}}.
\end{equation}
The two confidence components are computed as
\begin{equation}
\begin{aligned}
c_n^{\mathrm{ent}}
&=
1-\frac{H(q_n)}{\log 3},
\qquad
H(q_n)=-\sum_{r}q_n^{r}\log q_n^{r},\\
c_n^{\mathrm{scale}}
&=
1-\sigma(\log s_n),
\qquad
[\mu_n,\log s_n]=\phi_{\mathrm{dist}}(\ell_{n-1}).
\end{aligned}
\end{equation}
Here, \(c_n^{\mathrm{ent}}\) measures the sharpness of the coarse regime distribution, while \(c_n^{\mathrm{scale}}\) reflects the uncertainty of the continuous interval estimate. A larger \(c_n\) therefore indicates a sharper temporal-regime prediction and a more certain interval estimate, while a smaller \(c_n\) indicates higher temporal ambiguity.

TSCM is trained with a multi-task temporal objective:
\begin{equation}
\mathcal{L}_{\mathrm{time}}
= \mathcal{L}_{\mathrm{haz}}
+ \lambda_{\mathrm{dist}}\mathcal{L}_{\mathrm{dist}}
+ \lambda_{\mathrm{reg}}\mathcal{L}_{\mathrm{reg}}.
\label{eq:time-objective}
\end{equation}
Here, $\mathcal{L}_{\mathrm{haz}}$ is the discrete-time survival loss over fine-grained interval bins, $\mathcal{L}_{\mathrm{dist}}$ is a Gaussian NLL on $\log(1+\Delta_n)$, and $\mathcal{L}_{\mathrm{reg}}$ is the temporal-regime classification loss. After training, TSCM outputs \((z_n^{t}, q_n, c_n)\) at each generation step. These outputs together serve as a compact and uncertainty-aware temporal condition, guiding next-visit timing estimation and patient-specific clinical relation construction in the subsequent TRAM module.

\subsection{Temporal-Relational Adaptation Module}
\label{sec:TRAM}
The Temporal-Relational Adaptation Module (TRAM) explicitly models the intra-visit clinical structure. TRAM separates relation modeling into two complementary branches. The \textit{dynamic branch} captures temporal shifts in inter-variable correlations using the temporal outputs from TSCM, while the \textit{static branch} captures stable structure across data types directly from the patient history. The two branches are then merged through confidence-aware residual fusion and projected into patient-specific soft prefix tokens \(Z_n\).

\textbf{Dynamic Branch.}
The dynamic branch uses the current temporal state to modulate the relation representation. It takes as input the backbone history state \(h_{n-1}\), obtained by encoding the serialized patient history with the base generator described in Section~\ref{sec:generator}, together with the temporal embedding \(z_n^t\) from Section~\ref{sec:TSCM}.

Within this branch, the backbone history state is normalized and projected as \(\bar{h}_{n-1}=W_h\mathrm{LN}(h_{n-1})\), while the temporal embedding is mapped to FiLM modulation parameters \([\gamma_n,\beta_n]=W_tz_n^t\). The modulated history signal is then combined with the projected regime distribution:
\begin{equation}
z_n^{d}
=
\phi_d([\gamma_n \odot \bar{h}_{n-1} + \beta_n; W_q q_n]).
\end{equation}
where \(W_q\) and \(\phi_d(\cdot)\) are learnable projections. This branch captures temporal variation in patient-specific relation structure. Here, \(z_n^t\) provides continuous temporal conditioning, while \(q_n\) injects coarse regime information across short-, medium-, and long-interval temporal regimes. The confidence score \(c_n\) is further used in the subsequent fusion step to control how strongly this dynamic embedding contributes to the final relation representation.

\textbf{Static Branch.}
The static branch extracts time-agnostic clinical structure from the patient history, capturing stable relations across data types under different temporal regimes. It takes the same visit-level clinical code groups as TSCM as input, but uses separate embeddings and encoders for relation-oriented feature learning. We first obtain data-type-specific visit embeddings:
\begin{equation}
\bar{e}_i^{(k)}
=
\mathrm{MeanPool}\!\left(E_s^{(k)}(\mathcal{C}_i^{(k)})\right),
\qquad
k \in \mathcal{M}.
\end{equation}
Let \(\tau_i\) denote the visit type and \(e_i^{\mathrm{type}} = E_s^{\mathrm{type}}(\tau_i)\) its embedding. The pooled data-type embeddings and visit-type embedding are then passed through a shared visit encoder:
\begin{equation}
u_i
=
\phi_{\mathrm{visit}}^{s}
\!\left(
\left[
\operatorname*{Concat}_{k\in\mathcal{M}}\bar{e}_i^{(k)};
e_i^{\mathrm{type}}
\right]
\right).
\end{equation}
This produces the encoded visit sequence \(U_{1:n-1}=[u_1,\dots,u_{n-1}]\).

From this shared visit sequence, the static branch extracts three complementary summaries.

\emph{History summary.} Attention pooling summarizes the overall history:
\begin{equation}
z_n^{\mathrm{att}}
=
\sum_{i=1}^{n-1}
\omega_i u_i,
\qquad
\omega_i
=
\frac{\exp(a_g^\top u_i)}
{\sum_{j=1}^{n-1}\exp(a_g^\top u_j)}.
\end{equation}

\emph{Anchor summary.} Data-type-specific anchors are constructed from the historical visits:
\begin{equation}
a_n^{(k)}
=
\sum_{i=1}^{n-1}
\omega_i^{(k)}
W_k \bar{e}_i^{(k)},
\qquad
\omega_i^{(k)}
=
\frac{\exp(b_k^\top W_k \bar{e}_i^{(k)})}
{\sum_{j=1}^{n-1}\exp(b_k^\top W_k \bar{e}_j^{(k)})}.
\end{equation}
We summarize these anchors as
\begin{equation}
z_n^{\mathrm{anchor}}
=
\frac{1}{|\mathcal{M}|}
\sum_{k \in \mathcal{M}} a_n^{(k)}.
\end{equation}

\emph{Pairwise relation summary.} Interactions across data types are modeled explicitly. For each unordered data-type pair \((k,\ell)\), we construct
\begin{equation}
r_n^{(k,\ell)}
=
\phi_{\mathrm{pair}}
\!\left(
[a_n^{(k)};a_n^{(\ell)};
 |a_n^{(k)}-a_n^{(\ell)}|;
 a_n^{(k)} \odot a_n^{(\ell)}]
\right).
\end{equation}
Let \(\mathcal{P}=\{(k,\ell)\mid k,\ell\in\mathcal{M},\, k<\ell\}\) denote the set of unordered data-type pairs. We fuse these pairwise relation vectors into
\begin{equation}
z_n^{\mathrm{pair}}
=
\phi_{\mathrm{pair}}^{\mathrm{fuse}}
\!\left(
\operatorname*{Concat}_{(k,\ell)\in\mathcal{P}}
r_n^{(k,\ell)}
\right).
\end{equation}
Finally, the static relational embedding is formed by combining the attention-pooled history summary, the field-anchor summary, and the pairwise relation summary:
\begin{equation}
z_n^{s}
=
\phi_s
\!\left(
[z_n^{\mathrm{att}};z_n^{\mathrm{anchor}};z_n^{\mathrm{pair}}]
\right).
\end{equation}
This branch captures stable relation structure in patient history. It provides a time-invariant basis for dependencies across data types.

\textbf{Confidence-Aware Fusion.}
We use the confidence score \(c_n\) from TSCM to control how strongly the dynamic branch contributes to the fused relation representation:
\begin{equation}
\tilde{z}_n^{\mathrm{rel}}
=
\alpha_s \mathrm{LN}(z_n^{s})
+
\alpha_d c_n \mathrm{LN}(z_n^{d}),
\end{equation}
where \(\alpha_s\) and \(\alpha_d\) are learnable positive scaling factors, and \(c_n\) scales the contribution of the dynamic branch.

Finally, the fused relation representation is projected into patient-specific soft prefix tokens:
\begin{equation}
Z_n
=
\psi(\tilde{z}_n^{\mathrm{rel}})
\in \mathbb{R}^{m \times d},
\end{equation}
where \(m\) is the prefix length and \(d\) is the hidden dimension of the generator. These prefix tokens are then injected into the LoRA-adapted generator as described in Section~\ref{sec:generator}. Through this design, TRAM converts stable and temporally adaptive clinical relations into a lightweight conditioning signal for next-visit generation.

\subsection{LoRA-adapted LLM Generator}
\label{sec:generator}
% This section introduces the base generator of~\N. 
We design a parameter-efficient LoRA-adapted language model as the generator of SynEHR. As illustrated in Figure~\ref{fig:framework}, this module follows four steps: (i) input embedding, (ii) prefix injection, (iii) LoRA-based backbone encoding, and (iv) next-visit decoding.

\textbf{Input embedding.}
Let \(s_{1:n-1}\) denote the tokenized observed history. Its token embeddings are \(E(s_{1:n-1}) \in \mathbb{R}^{L \times d}\), where \(L\) is the sequence length and \(d\) is the hidden dimension. From the same input, the generator derives a compact history representation \(h_{n-1}=f_{LM}(s_{1:n-1})\), which is passed to the downstream adaptation modules.

\textbf{Prefix injection.}
In the LoRA-adapted LLM generator, the patient-specific soft prefix \(Z_n \in \mathbb{R}^{m \times d}\) generated by TRAM provides the key mechanism for integrating inter-visit temporal evolution and intra-visit clinical structure at the conditioning level. We prepend it to the token embedding sequence as \(\tilde{E}(s_{1:n-1})=[Z_n;E(s_{1:n-1})]\), where \([\,;\,]\) denotes concatenation along the token dimension. In this way, the injected soft prefix provides patient-specific conditioning context while leaving the discrete history tokens unchanged.

\textbf{LoRA-adapted generator.}
The augmented embedding sequence \(\widetilde{E}(s_{1:n-1})\) is processed by the LoRA-adapted causal transformer. LoRA adds low-rank trainable updates to the pretrained LLM while freezing its original parameters, enabling \textbf{parameter-efficient adaptation} without changing the generator architecture. In staged training, this generator is first optimized with the autoregressive next-visit objective and then fixed when training TSCM and TRAM.

\textbf{Next-visit decoding.}
Given the augmented embedding sequence \(\widetilde{E}(s_{1:n-1})\), the LoRA-adapted generator produces the next visit in a structured form, including the inter-visit interval and the corresponding clinical codes across data types. If the patient trajectory terminates, the generator outputs an end token.
% instead of a next visit.

During training, the ground-truth next visit \(x_n\) is serialized into a target token sequence \(y_n = (y_{n,1}, \dots, y_{n,T_n})\). The autoregressive training objective is
\begin{equation}
\mathcal{L}_{\mathrm{AR}}
= - {\textstyle\sum_{n=2}^{N}\sum_{t=1}^{T_n}}
\log p_{\theta}\!\left(y_{n,t} \mid y_{n,<t}, \widetilde{E}(s_{1:n-1})\right),
\label{eq:ar-objective}
\end{equation}
where \(p_{\theta}\) denotes the next-token distribution parameterized by the LoRA-adapted generator. In the first training stage, the patient-specific soft prefix is absent, so \(\widetilde{E}(s_{1:n-1})\) reduces to the standard embedding sequence of the serialized history, and the objective becomes standard next-visit autoregressive training. After the patient-specific soft prefix is introduced, next-visit generation is conditioned on patient-specific temporal and structural content, enabling adaptive next-visit generation.

\subsection{Training and Inference Strategy}
\label{sec:training_inference}

\N is trained in three stages to progressively learn next-visit generation, temporal conditioning, and prefix adaptation:

\textbf{Stage 1.} We train the base generator with the autoregressive next-visit generation objective \(\mathcal{L}_{\mathrm{AR}}\) from Eq.~\eqref{eq:ar-objective}. This stage equips the LoRA-adapted generator with the basic ability to synthesize the next visit from the observed history.

\textbf{Stage 2.} We freeze the base generator and train TSCM with the temporal objective \(\mathcal{L}_{\mathrm{time}}\) from Eq.~\eqref{eq:time-objective} on step-wise trajectory content. This stage learns a temporal state that summarizes inter-visit dynamics and uncertainty.

\textbf{Stage 3.} We freeze both the base generator and TSCM, inject the learned patient-specific soft prefixes into the input, and compute \(\mathcal{L}_{\mathrm{AR}}\) from Eq.~\eqref{eq:ar-objective} through the full prefix-conditioned generator. 
This stage learns relation-aware prefix conditioning that adapts the frozen generator to temporal and intra-visit clinical structure.

During inference, SynEHR performs visit-level autoregressive rollout. Given the current history \(s_{1:n-1}\), it constructs the patient-specific soft prefix \(Z_n\) following the procedure described above and feeds the prefix-augmented input into the LoRA-adapted generator to synthesize the next visit \(x_n\). Once \(x_n\) is generated, it is added to the patient trajectory to form the updated history \(s_{1:n}\), which is then used to generate the following visit. This iterative process continues until the generator produces an end-of-sequence token.

\section{Evaluation}

\subsection{Experimental Setup}

\subsubsection{Datasets}
We evaluate~\N on two publicly available critical-care EHR datasets, MIMIC-III~\cite{johnson2016mimiciii, PhysioNet-mimiciii-1.4} and MIMIC-IV~\cite{johnson2023mimiciv, PhysioNet-mimiciv-2.2}, collected from Beth Israel Deaconess Medical Center and covering admissions from 2001--2012 and 2008--2019, respectively. For both datasets, we retain patients with at least three visits. Each visit contains four clinical data types, including diagnoses, procedures, medications, and laboratory items, together with visit-level attributes. We also include demographic information, such as age, sex, and race, as static patient features. 
Dataset statistics are summarized in Table~\ref{tab:dataset_stats}.

\begin{table}[t]\small
\centering
\caption{Statistics of used datasets.}
\renewcommand{\arraystretch}{1.15}
\setlength{\tabcolsep}{8pt}
\label{tab:dataset_stats}
\begin{tabular}{l|r||l|r}
\hline
\multicolumn{2}{c||}{\textbf{MIMIC-III}} & \multicolumn{2}{c}{\textbf{MIMIC-IV}} \\
\hline
Total Patients  & 1,795 & Total Patients  & 56,028 \\
Total Visits    & 7,274 & Total Visits    & 334,391 \\
Diagnosis       & 3,366 & Diagnosis       & 24,062 \\
Medications     & 2,363 & Medications     & 4,479 \\
Lab Items       & 633   & Lab Items       & 893 \\
Procedure       & 1,019 & Procedure       & 12,218 \\
\hline
\end{tabular}
\end{table}

\subsubsection{Evaluation Metrics}
\label{sec:metrics}

We evaluate~\N from three complementary perspectives: fidelity, privacy, and downstream utility.
\begin{itemize}[leftmargin=*]
    \item \textbf{Fidelity.}
    We assess whether synthetic data preserves the temporal, clinical, and cross-data-type patterns of real longitudinal EHRs. We use Longitudinal Imputation Perplexity (LPL) to measure temporal consistency within each data type, and Cross-modality Imputation Perplexity (MPL) to measure dependency preservation across data types within the same visit. Concretely, LPL scores target codes using only the preceding visit history, whereas MPL additionally conditions on the other data types within the same visit. In our setting, the modalities correspond to clinical data types, including diagnoses, procedures, medications, and laboratory items. For LLM-based models, both metrics are computed from token-level log-probabilities under structured prompts. In both cases, the normalization counts only clinical-code tokens and excludes structural JSON tokens such as braces, field names, commas, and delimiters. Lower LPL and MPL indicate better fidelity. We also report Jensen-Shannon Divergence (JSD) as a supplementary distribution-level metric.

    \item \textbf{Privacy.}
    We evaluate privacy using Presence Disclosure Sensitivity (PD), which measures the risk of linking generated samples back to real training patients under a similarity-based attack protocol. Lower PD indicates lower disclosure risk and stronger privacy protection.

    \item \textbf{Downstream Utility.}
    We adopt a train-on-synthetic, test-on-real protocol to evaluate whether synthetic data preserves task-relevant clinical signals. For risk prediction using multiple data types, we report the Area Under the Receiver Operating Characteristic Curve (AUROC) and the Area Under the Precision-Recall Curve (AUPRC): AUROC evaluates overall ranking quality across decision thresholds, while AUPRC is more informative for positive-case detection under class imbalance. For time interval prediction, we report Macro-F1, Mean Absolute Error (MAE), Negative Log-Likelihood (NLL), Brier score, and Expected Calibration Error (ECE). Macro-F1 evaluates balanced regime discrimination across short-, medium-, and long-interval classes; MAE measures point-estimation error in days; NLL evaluates how much probability mass is assigned to the true interval regime; the Brier score reflects the accuracy and confidence quality of predicted probabilities; and ECE measures probability calibration.
    For interval-regime prediction with $K$ classes, the Brier score is
    % defined as
    \begin{equation}
    \mathrm{BS}
    =
    \frac{1}{N}\sum_{i=1}^{N}\sum_{k=1}^{K}
    \left(p_{i,k}-y_{i,k}\right)^2,
    \end{equation}
    where $p_{i,k}$ is the predicted probability for class $k$ on sample $i$, and $y_{i,k}\in\{0,1\}$ is the one-hot ground-truth label. We compute ECE by partitioning predictions into $M$ confidence bins:
    \begin{equation}
    \mathrm{ECE}
    =
    \sum_{m=1}^{M}\frac{|B_m|}{N}
    \left|
    \mathrm{acc}(B_m)-\mathrm{conf}(B_m)
    \right|,
    \end{equation}
    where $B_m$ is the set of samples whose predicted confidence falls into bin $m$, $\mathrm{acc}(B_m)$ is the empirical accuracy of that bin, and $\mathrm{conf}(B_m)$ is the average predicted confidence. In our experiments, we use $M=10$ equal-width confidence bins.
\end{itemize}

\subsubsection{Baselines}
Our baseline models include recurrent predictors such as
MLP~\cite{hochreiter1997lstm} and
BEHRT~\cite{li2020behrt}, GAN-based models including
medGAN~\cite{choi2017medgan} and
SynTEG~\cite{zhang2021synteg}, and VAE-based models including
EVA~\cite{biswal2021eva} and
TWIN~\cite{das2023twin}. We further compare against
diffusion-based approaches, including
TabDDPM~\cite{kotelnikov2023tabddpm},
MedDiff~\cite{he2023meddiff},
ScoEHR~\cite{naseer2023scoehr}, and
EHRPD~\cite{zhong2024ehrpd}, as well as language-model-based
approaches, including PromptEHR~\cite{wang2022promptehr},
HALO~\cite{theodorou2023halo},
HiSGT~\cite{zhou2025hisgt}, and
EHR2Path~\cite{pellegrini2025ehr2path}. Their key characteristics
are summarized below:
\begin{itemize}[leftmargin=*, nosep, partopsep=4pt]
    \item \textbf{MLP}~\cite{hochreiter1997lstm} adopts an LSTM encoder followed by an MLP prediction head to model visit-to-visit dependencies.
    \item \textbf{BEHRT}~\cite{li2020behrt} is a non-generative transformer-based method that learns contextualized representations from structured longitudinal EHRs for clinical prediction.
    \item \textbf{medGAN}~\cite{choi2017medgan} generates high-dimensional discrete patient records with adversarial training.
    \item \textbf{SynTEG}~\cite{zhang2021synteg} combines sequential temporal modeling with GAN-based visit generation.
    \item \textbf{EVA}~\cite{biswal2021eva} learns latent visit representations for EHR synthesis.
    \item \textbf{TWIN}~\cite{das2023twin} uses a VAE-style architecture with decoders for current- and next-visit code generation to better capture dependencies across data types and over time.
    \item \textbf{TabDDPM}~\cite{kotelnikov2023tabddpm} is a diffusion model for tabular data that we adapt to visit-level EHR generation.
    \item \textbf{MedDiff}~\cite{he2023meddiff} employs an efficient diffusion-based synthesis procedure to aggregated discrete EHR features.
    \item \textbf{ScoEHR}~\cite{naseer2023scoehr} builds on a continuous-time diffusion method to model irregular temporal progression in patient records.
    \item \textbf{EHRPD}~\cite{zhong2024ehrpd} is a predictive diffusion model that generates the next visit together with the corresponding time interval.
    \item \textbf{PromptEHR}~\cite{wang2022promptehr} formulates EHR generation as prompt-conditioned sequence generation with a pre-trained BART model.
    \item \textbf{HALO}~\cite{theodorou2023halo} employs a hierarchical autoregressive transformer architecture to jointly model code- and visit-level dependencies in high-dimensional longitudinal EHR sequences.
    \item \textbf{HiSGT}~\cite{zhou2025hisgt} is a transformer based approach for synthetic EHR generation that leverages code hierarchies and clinical semantic embeddings to improve clinical fidelity.
    \item \textbf{EHR2Path}~\cite{pellegrini2025ehr2path} is a pathway modeling baseline over multiple clinical data types, designed for patient-pathway forecasting and simulation and providing a strong long-context comparator for longitudinal trajectory modeling.
\end{itemize}

\begin{table*}[t]
\centering
\caption{Code-level fidelity comparison. Lower is better. \N-L3.1,\N-Q2.5, and \N-Q3 denote~\N with LLaMA-3.1, Qwen2.5, and Qwen3 backbones, respectively. Best values are shown in bold, and second-best values are underlined.}
\scriptsize
\renewcommand{\arraystretch}{1.12}
\setlength{\tabcolsep}{3.2pt}
\resizebox{\textwidth}{!}{
\begin{tabular}{c|c|c|c|c|c|c|c|c|c|c|c|c|c|c|c|c|c|c|c}
\specialrule{0.5pt}{0pt}{0pt}
\textbf{Dataset} & \textbf{Data Type} & \textbf{Metric} & \textbf{MLP} & \textbf{BEHRT} & \textbf{medGAN} & \textbf{SynTEG} & \textbf{EVA} & \textbf{TWIN} & \textbf{TabDDPM} & \textbf{MedDiff} & \textbf{ScoEHR} & \textbf{EHRPD} & \textbf{PromptEHR} & \textbf{HALO} & \textbf{HiSGT} & \textbf{EHR2Path} & \textbf{\N-L3.1} & \textbf{\N-Q2.5} & \textbf{\N-Q3} \\
\hline
\multirow{8}{*}{\rotatebox{90}{\textbf{MIMIC-III}}}
& \multirow{2}{*}{Diagnosis}  & LPL & 318.74 & 96.42 & 236.48 & 35.76 & 28.94 & 26.11 & 111.27 & 651.38 & 672.44 & 16.42 & 72.84 & 88.57 & 39.86 & 15.73 & 16.18 & \underline{15.61} & \textbf{15.34} \\
&            & MPL & 305.86 & 91.35 & 224.17 & 33.94 & 27.48 & 24.96 & 103.84 & 629.57 & 648.63 & \underline{15.51} & 68.71 & 81.42 & 37.24 & 15.63 & 15.88 & 15.54 & \textbf{15.46} \\
\cline{2-20}
& \multirow{2}{*}{Medication} & LPL & 541.28 & 148.73 & 394.67 & 74.68 & 41.52 & 39.36 & 182.54 & 918.63 & 921.46 & 20.46 & 88.34 & 96.18 & 63.92 & 18.96 & 19.48 & \underline{18.81} & \textbf{18.39} \\
&            & MPL & 508.46 & 142.51 & 376.42 & 71.93 & 39.68 & 38.12 & 169.87 & 887.14 & 896.85 & \textbf{18.74} & 81.57 & 89.33 & 60.85 & 18.91 & 19.02 & 18.83 & \underline{18.79} \\
\cline{2-20}
& \multirow{2}{*}{Lab Item}   & LPL & 162.47 & 48.37 & 79.84 & 24.96 & 18.74 & 16.91 & 56.48 & 401.92 & 419.35 & 15.08 & 39.46 & 84.72 & 22.41 & \textbf{14.51} & 14.88 & 14.63 & \underline{14.57} \\
&            & MPL & 154.32 & 45.82 & 74.19 & 23.41 & 17.82 & 16.36 & 51.27 & 388.16 & 402.84 & \textbf{14.22} & 35.78 & 76.39 & 21.08 & 14.37 & 14.58 & \underline{14.31} & 14.41 \\
\cline{2-20}
& \multirow{2}{*}{Procedure}  & LPL & 284.16 & 82.65 & 229.57 & 43.18 & 24.95 & 20.27 & 96.12 & 459.28 & 473.66 & 15.11 & 31.84 & 29.42 & 33.57 & 14.79 & 15.02 & \underline{14.73} & \textbf{14.66} \\
&            & MPL & 269.47 & 78.51 & 218.93 & 41.62 & 23.71 & 19.84 & 91.54 & 446.95 & 460.28 & \textbf{14.99} & 29.36 & 27.85 & 31.44 & 15.06 & 15.15 & \underline{15.03} & 15.07 \\
\hline
\multirow{8}{*}{\rotatebox{90}{\textbf{MIMIC-IV}}}
& \multirow{2}{*}{Diagnosis}  & LPL & 350.61 & 107.03 & 257.76 & 40.05 & 31.83 & 28.46 & 122.40 & 703.49 & 732.96 & 18.23 & 80.85 & 99.20 & 43.45 & 17.20 & 17.08 & \underline{16.92} & \textbf{16.79} \\
&            & MPL & 342.56 & 103.23 & 246.59 & 38.69 & 30.50 & 27.46 & 116.30 & 686.23 & 713.49 & 17.84 & 76.96 & 92.82 & 40.96 & 17.30 & 17.14 & \underline{16.97} & \textbf{16.84} \\
\cline{2-20}
& \multirow{2}{*}{Medication} & LPL & 384.31 & 107.09 & 276.27 & 54.52 & 29.48 & 27.55 & 131.43 & 633.85 & 635.81 & 14.53 & 63.60 & 70.21 & 44.74 & 13.17 & 13.10 & \underline{12.99} & \textbf{12.92} \\
&            & MPL & 366.09 & 104.03 & 267.26 & 53.23 & 28.57 & 27.07 & 124.01 & 621.00 & 627.79 & 13.49 & 59.55 & 66.10 & 43.20 & 13.34 & 13.27 & \underline{13.18} & \textbf{13.11} \\
\cline{2-20}
& \multirow{2}{*}{Lab Item}   & LPL & 131.60 & 39.66 & 63.87 & 20.72 & 15.18 & 13.53 & 46.31 & 317.52 & 331.29 & 12.52 & 32.36 & 71.16 & 17.93 & 11.59 & 11.63 & \textbf{11.54} & \underline{11.56} \\
&            & MPL & 123.46 & 37.11 & 58.61 & 19.20 & 14.26 & 12.92 & 41.53 & 302.76 & 314.22 & 11.94 & 28.98 & 63.40 & 16.65 & \underline{11.61} & 11.68 & \textbf{11.60} & 11.63 \\
\cline{2-20}
& \multirow{2}{*}{Procedure}  & LPL & 318.26 & 93.39 & 254.82 & 49.66 & 27.94 & 22.50 & 108.62 & 505.21 & 525.76 & 16.92 & 36.30 & 33.83 & 37.26 & 16.56 & 16.46 & \underline{16.33} & \textbf{16.24} \\
&            & MPL & 307.20 & 90.29 & 247.39 & 48.70 & 27.03 & 22.42 & 105.27 & 500.58 & 520.12 & 17.69 & 34.06 & 32.58 & 35.84 & \underline{17.41} & 17.55 & 17.44 & \textbf{17.36} \\
\specialrule{0.5pt}{0pt}{0pt}
\end{tabular}
}
\label{tab:fidelity_main}
\end{table*}

\vspace{-5pt}

\begin{table*}[t]
\centering
\caption{JSD between real and synthetic data distributions. Lower is better.}
\scriptsize
\renewcommand{\arraystretch}{1.12}
\setlength{\tabcolsep}{3.2pt}
\resizebox{\textwidth}{!}{
\begin{tabular}{c|c|c|c|c|c|c|c|c|c|c|c|c|c|c|c|c|c|c}
\specialrule{0.5pt}{0pt}{0pt}
\textbf{Dataset} & \textbf{Statistic} & \textbf{MLP} & \textbf{BEHRT} & \textbf{medGAN} & \textbf{SynTEG} & \textbf{EVA} & \textbf{TWIN} & \textbf{TabDDPM} & \textbf{MedDiff} & \textbf{ScoEHR} & \textbf{EHRPD} & \textbf{PromptEHR} & \textbf{HALO} & \textbf{HiSGT} & \textbf{EHR2Path} & \textbf{\N-L3.1} & \textbf{\N-Q2.5} & \textbf{\N-Q3} \\
\hline
\multirow{6}{*}{\rotatebox{90}{\textbf{MIMIC-III}}}
& Diagnosis Code Freq.  & 0.346 & 0.186 & 0.337 & 0.378 & 0.274 & 0.290 & 0.152 & 0.134 & 0.180 & 0.228 & 0.224 & 0.361 & 0.229 & \textbf{0.042} & \underline{0.079} & 0.146 & 0.129 \\
\cline{2-19}
& Medication Code Freq. & 0.379 & 0.201 & 0.415 & 0.291 & 0.334 & 0.241 & 0.192 & 0.190 & 0.231 & 0.255 & 0.229 & 0.329 & 0.234 & \textbf{0.138} & \underline{0.176} & 0.187 & 0.181 \\
\cline{2-19}
& Lab Item Code Freq.   & 0.038 & 0.095 & 0.047 & 0.047 & \underline{0.014} & 0.152 & 0.113 & 0.201 & 0.241 & 0.248 & 0.162 & 0.057 & 0.065 & 0.024 & \textbf{0.013} & 0.021 & 0.023 \\
\cline{2-19}
& Procedure Code Freq.  & 0.430 & 0.301 & 0.298 & 0.494 & 0.283 & 0.360 & 0.148 & 0.199 & 0.241 & 0.262 & 0.258 & 0.401 & 0.331 & \underline{0.111} & \textbf{0.106} & 0.168 & 0.141 \\
\cline{2-19}
& Visit Length          & 0.193 & 0.266 & 0.662 & 0.632 & 0.116 & 0.268 & 0.046 & 0.425 & 0.371 & 0.559 & 0.224 & 0.662 & 0.040 & 0.640 & \textbf{0.017} & 0.034 & \underline{0.029} \\
\cline{2-19}
& Inter-visit Time      & 0.250 & 0.227 & 0.274 & 0.332 & 0.371 & 0.171 & \underline{0.055} & 0.071 & 0.090 & 0.193 & 0.243 & 0.274 & 0.159 & 0.271 & \textbf{0.049} & 0.068 & 0.061 \\
\hline
\multirow{6}{*}{\rotatebox{90}{\textbf{MIMIC-IV}}}
& Diagnosis Code Freq.  & 0.113 & 0.088 & 0.439 & 0.338 & 0.221 & 0.297 & 0.218 & \textbf{0.014} & 0.036 & 0.051 & 0.104 & 0.425 & 0.103 & 0.039 & \underline{0.026} & 0.049 & 0.038 \\
\cline{2-19}
& Medication Code Freq. & 0.220 & 0.131 & 0.404 & 0.248 & 0.329 & 0.261 & 0.213 & \textbf{0.038} & 0.098 & 0.102 & 0.147 & 0.381 & 0.145 & 0.046 & \underline{0.042} & 0.062 & 0.053 \\
\cline{2-19}
& Lab Item Code Freq.   & 0.054 & 0.059 & 0.143 & 0.142 & 0.040 & 0.117 & 0.142 & 0.221 & 0.286 & 0.276 & 0.065 & 0.186 & 0.058 & \underline{0.020} & \textbf{0.018} & 0.031 & 0.025 \\
\cline{2-19}
& Procedure Code Freq.  & 0.515 & 0.383 & 0.418 & 0.507 & 0.475 & 0.345 & 0.263 & \underline{0.034} & 0.056 & 0.054 & 0.453 & 0.447 & 0.325 & 0.115 & \textbf{0.032} & 0.058 & 0.046 \\
\cline{2-19}
& Visit Length          & 0.076 & 0.043 & 0.102 & 0.100 & 0.075 & 0.053 & \textbf{0.009} & 0.044 & 0.085 & 0.110 & 0.062 & 0.102 & 0.037 & 0.046 & \underline{0.021} & 0.038 & 0.030 \\
\cline{2-19}
& Inter-visit Time      & 0.119 & 0.118 & 0.268 & 0.275 & 0.134 & 0.143 & 0.163 & \underline{0.036} & 0.070 & 0.146 & 0.133 & 0.268 & 0.098 & 0.056 & \textbf{0.031} & 0.056 & 0.045 \\
\specialrule{0.5pt}{0pt}{0pt}
\end{tabular}
}
\label{tab:jsd_appendix}
\end{table*}

\subsubsection{Implementation Details}
We implement all models in PyTorch and train them on an NVIDIA B200 GPU. We evaluate three instruction-tuned backbones, LLaMA-3.1-8B-Instruct~\cite{llama31_8b_instruct}, Qwen2.5-7B-Instruct~\cite{qwen2024qwen25}, and Qwen3-4B-Instruct-2507~\cite{qwen2025qwen3}, all in \texttt{bfloat16} precision. Each backbone is adapted with LoRA\cite{hu2022lora} using rank $64$, $\alpha=128$, and dropout $0.05$. For fine-tuning, we use AdamW with a learning rate of $10^{-4}$, cosine decay, a warm-up ratio of $0.03$, a batch size of $32$, and 10 training epochs. The TSCM and TRAM are trained in separate stages using AdamW, with learning rates of $10^{-3}$ and $10^{-4}$, respectively. All random seeds are fixed to 42.

%, and cosine-based scheduling where applicable
\vspace{-10pt}
\subsection{Fidelity Performance}

% We first evaluate whether~\N can faithfully reproduce the temporal, clinical, and cross-data-type patterns of real longitudinal EHR trajectories. 
Table~\ref{tab:fidelity_main} reports the main fidelity results on MIMIC-III and MIMIC-IV in terms of Longitudinal Imputation Perplexity (LPL) and Cross-modality Imputation Perplexity (MPL).

Table~\ref{tab:fidelity_main} shows that SynEHR improves code-level fidelity on both MIMIC-III and MIMIC-IV. On MIMIC-III, SynEHR-Q3 reduces diagnosis, medication, and procedure LPL by 2.5\%, 3.0\%, and 0.9\%, respectively, relative to the strongest non-SynEHR baseline. It also stays within 0.4\% of the best lab-item LPL and lowers diagnosis MPL by 0.3\%. On MIMIC-IV, the same pattern holds: SynEHR-Q3 reduces diagnosis, medication, and procedure LPL by 2.4\%, 1.9\%, and 1.9\%, and also lowers diagnosis and medication MPL by 2.7\% and 1.7\%. For lab items, SynEHR-Q2.5 slightly outperforms the strongest non-SynEHR baseline on both LPL and MPL.

Table~\ref{tab:jsd_appendix} shows that ~\N achieves strong distribution-level fidelity, with clear gains on several temporal and structural statistics. On MIMIC-III, SynEHR-L3.1 reduces JSD on lab-item frequency, procedure frequency, visit length, and inter-visit time by 7.1\%, 4.5\%, 57.5\%, and 10.9\%, respectively. On MIMIC-IV, compared with EHR2Path, it reduces JSD by 33.3\% on diagnosis frequency, 8.7\% on medication frequency, 10.0\% on lab-item frequency, 72.2\% on procedure frequency, 54.3\% on visit length, and 44.6\% on inter-visit time. The largest gains appear on visit length and inter-visit time, suggesting that explicit temporal conditioning helps preserve global clinical structure as well as next-visit code predictability.

\begin{figure}[h]
  \centering
  \includegraphics[width=0.98\linewidth]{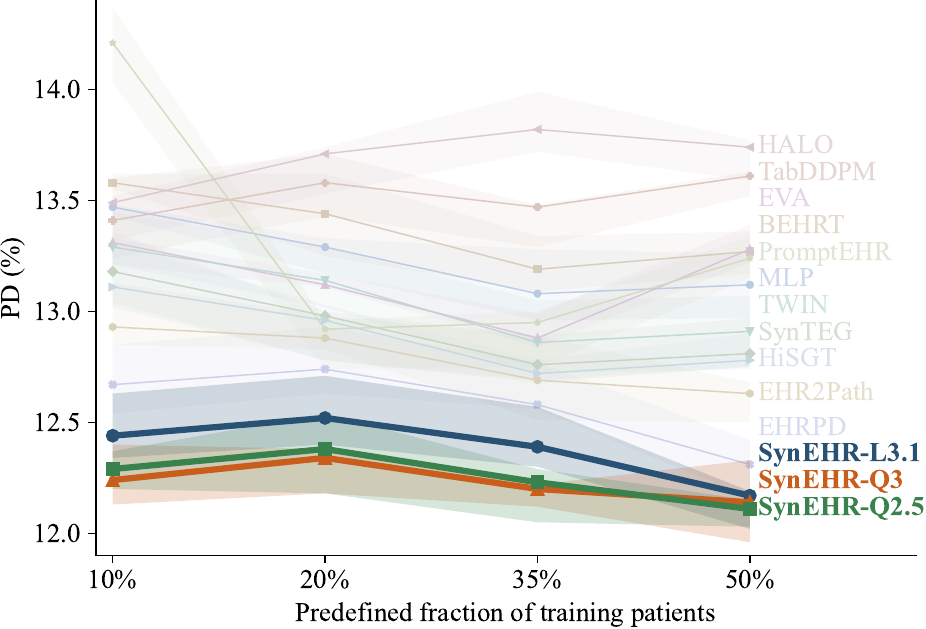}
  \caption{Privacy assessment on MIMIC-III. Lower is better.}
  \label{fig:privacy_pd}
\end{figure}

%Due to limited space, we omit MedDiff, medGAN, and ScoEHR from this view.

\subsection{Privacy Assessment}

% We next evaluate whether the synthetic trajectories generated by~\N leak patient-specific information from the training set. 
We use Presence Disclosure Sensitivity (PD) as the privacy metric and consider four attacker-knowledge settings, where the adversary is assumed to know 10\%, 20\%, 35\%, or 50\% of the real training patients in advance. Lower PD indicates lower disclosure risk.
% We compare~\N with the same baselines used in the fidelity assessment. 
All results are averaged over five random runs, with standard deviations reported as confidence intervals. Figure~\ref{fig:privacy_pd} visualizes the methods within the main PD range for readability; high-PD baselines are reported in Table~\ref{tab:privacy_pd}. The results show that SynEHR consistently lowers disclosure risk across all four settings. Compared with the strongest non-\N baseline, EHRPD, the best~\N variant reduces PD by 3.4\%, 3.1\%, 3.0\%, and 1.6\% under the four attacker-knowledge settings, respectively. This consistent margin suggests that the fidelity and utility gains of~\N do not come from memorizing training patients.

\begin{table}[t]
\centering
\caption{PD Comparisons.}
\footnotesize
\renewcommand{\arraystretch}{1.12}
\setlength{\tabcolsep}{3.2pt}
\begin{tabular}{l|c|c|c|c}
\specialrule{0.5pt}{0pt}{0pt}
\textbf{Approach} & \textbf{10\%} & \textbf{20\%} & \textbf{35\%} & \textbf{50\%} \\
\hline
MLP & \shortstack[c]{13.47\\{\tiny[13.43, 13.48]}} & \shortstack[c]{13.29\\{\tiny[13.17, 13.33]}} & \shortstack[c]{13.08\\{\tiny[12.96, 13.28]}} & \shortstack[c]{13.12\\{\tiny[12.97, 13.26]}} \\
BEHRT & \shortstack[c]{13.58\\{\tiny[13.55, 13.62]}} & \shortstack[c]{13.44\\{\tiny[13.25, 13.62]}} & \shortstack[c]{13.19\\{\tiny[13.09, 13.34]}} & \shortstack[c]{13.27\\{\tiny[13.17, 13.36]}} \\
medGAN & \shortstack[c]{16.94\\{\tiny[16.78, 17.04]}} & \shortstack[c]{17.11\\{\tiny[16.92, 17.13]}} & \shortstack[c]{17.43\\{\tiny[17.33, 17.55]}} & \shortstack[c]{17.67\\{\tiny[17.47, 17.73]}} \\
SynTEG & \shortstack[c]{13.18\\{\tiny[13.02, 13.36]}} & \shortstack[c]{12.98\\{\tiny[12.78, 13.00]}} & \shortstack[c]{12.76\\{\tiny[12.68, 12.89]}} & \shortstack[c]{12.81\\{\tiny[12.75, 12.97]}} \\
EVA & \shortstack[c]{13.31\\{\tiny[13.21, 13.33]}} & \shortstack[c]{13.12\\{\tiny[13.11, 13.17]}} & \shortstack[c]{12.88\\{\tiny[12.84, 12.99]}} & \shortstack[c]{13.28\\{\tiny[13.22, 13.39]}} \\
TWIN & \shortstack[c]{13.29\\{\tiny[13.21, 13.45]}} & \shortstack[c]{13.14\\{\tiny[13.02, 13.29]}} & \shortstack[c]{12.86\\{\tiny[12.72, 13.05]}} & \shortstack[c]{12.91\\{\tiny[12.74, 13.07]}} \\
TabDDPM & \shortstack[c]{13.41\\{\tiny[13.24, 13.60]}} & \shortstack[c]{13.58\\{\tiny[13.43, 13.71]}} & \shortstack[c]{13.47\\{\tiny[13.29, 13.48]}} & \shortstack[c]{13.61\\{\tiny[13.52, 13.63]}} \\
MedDiff & \shortstack[c]{14.88\\{\tiny[14.69, 14.94]}} & \shortstack[c]{16.79\\{\tiny[16.60, 16.80]}} & \shortstack[c]{18.11\\{\tiny[17.97, 18.23]}} & \shortstack[c]{18.92\\{\tiny[18.79, 18.98]}} \\
ScoEHR & \shortstack[c]{14.06\\{\tiny[14.04, 14.13]}} & \shortstack[c]{15.37\\{\tiny[15.30, 15.55]}} & \shortstack[c]{16.29\\{\tiny[16.24, 16.37]}} & \shortstack[c]{16.73\\{\tiny[16.70, 16.74]}} \\
EHRPD & \shortstack[c]{12.67\\{\tiny[12.54, 12.85]}} & \shortstack[c]{12.74\\{\tiny[12.63, 12.93]}} & \shortstack[c]{12.58\\{\tiny[12.56, 12.77]}} & \shortstack[c]{12.31\\{\tiny[12.16, 12.42]}} \\
PromptEHR & \shortstack[c]{14.21\\{\tiny[14.03, 14.37]}} & \shortstack[c]{12.92\\{\tiny[12.74, 12.96]}} & \shortstack[c]{12.95\\{\tiny[12.75, 13.00]}} & \shortstack[c]{13.24\\{\tiny[13.15, 13.34]}} \\
HALO & \shortstack[c]{13.49\\{\tiny[13.32, 13.52]}} & \shortstack[c]{13.71\\{\tiny[13.53, 13.74]}} & \shortstack[c]{13.82\\{\tiny[13.72, 13.99]}} & \shortstack[c]{13.74\\{\tiny[13.59, 13.77]}} \\
HiSGT & \shortstack[c]{13.11\\{\tiny[13.04, 13.31]}} & \shortstack[c]{12.96\\{\tiny[12.78, 13.03]}} & \shortstack[c]{12.72\\{\tiny[12.69, 12.78]}} & \shortstack[c]{12.78\\{\tiny[12.75, 12.90]}} \\
EHR2Path & \shortstack[c]{12.93\\{\tiny[12.85, 13.13]}} & \shortstack[c]{12.88\\{\tiny[12.83, 12.98]}} & \shortstack[c]{12.69\\{\tiny[12.51, 12.89]}} & \shortstack[c]{12.63\\{\tiny[12.50, 12.68]}} \\
\hline
\N-L3.1 & \shortstack[c]{12.44\\{\tiny[12.34, 12.63]}} & \shortstack[c]{12.52\\{\tiny[12.40, 12.71]}} & \shortstack[c]{12.39\\{\tiny[12.30, 12.57]}} & \shortstack[c]{12.17\\{\tiny[12.02, 12.19]}} \\
\N-Q2.5 & \shortstack[c]{\underline{12.29}\\{\tiny[12.20, 12.37]}} & \shortstack[c]{\underline{12.38}\\{\tiny[12.18, 12.54]}} & \shortstack[c]{\underline{12.23}\\{\tiny[12.05, 12.29]}} & \shortstack[c]{\textbf{12.11}\\{\tiny[12.03, 12.15]}} \\
\N-Q3 & \shortstack[c]{\textbf{12.24}\\{\tiny[12.13, 12.40]}} & \shortstack[c]{\textbf{12.34}\\{\tiny[12.18, 12.38]}} & \shortstack[c]{\textbf{12.20}\\{\tiny[12.12, 12.22]}} & \shortstack[c]{\underline{12.14}\\{\tiny[11.96, 12.33]}} \\
\specialrule{0.5pt}{0pt}{0pt}
\end{tabular}
\label{tab:privacy_pd}
\end{table}

%Each entry reports the mean PD after 5 random samplings of the compromised patient set, along with the standard deviation in brackets.

\subsection{Downstream Utility}

We further evaluate whether the synthetic data generated by~\N preserve task-relevant signals for downstream clinical modeling under a train-on-synthetic, test-on-real protocol. We consider two tasks: risk prediction using multiple data types and time interval prediction, which assess clinical structure across data types and temporal dynamics, respectively.

\subsubsection{Risk Prediction Using Multiple Data Types}

For risk prediction using multiple data types, we align the four data-type-specific streams at the visit level and construct prediction labels for acute respiratory failure (ARF), shock, and mortality by adapting the outcome definitions from FIDDLE~\cite{tang2020fiddle} to our cohort. We report AUROC and AUPRC for these three tasks.

\begin{table}[t]
\centering
\caption{Risk prediction results using multiple data types. Each entry is reported as AUROC / AUPRC. The Avg. column averages the three task-specific scores under the same metric. }
\footnotesize
\renewcommand{\arraystretch}{1.12}
\setlength{\tabcolsep}{4.8pt}
\begin{tabular}{l|c|c|c|c}
\specialrule{0.5pt}{0pt}{0pt}
\textbf{Model} & \textbf{ARF} & \textbf{Shock} & \textbf{Mortality} & \textbf{Avg.} \\
\hline
MLP & 0.690 / 0.318 & 0.652 / 0.236 & 0.701 / 0.207 & 0.681 / 0.254 \\
BEHRT & 0.716 / 0.347 & 0.674 / 0.258 & 0.732 / 0.230 & 0.707 / 0.278 \\
medGAN & 0.648 / 0.283 & 0.615 / 0.198 & 0.666 / 0.178 & 0.643 / 0.220 \\
SynTEG & 0.673 / 0.302 & 0.637 / 0.216 & 0.688 / 0.190 & 0.666 / 0.236 \\
EVA & 0.692 / 0.324 & 0.655 / 0.236 & 0.708 / 0.215 & 0.685 / 0.258 \\
TWIN & 0.706 / 0.337 & 0.671 / 0.258 & 0.723 / 0.225 & 0.700 / 0.273 \\
TabDDPM & 0.731 / 0.366 & 0.687 / 0.272 & 0.747 / 0.259 & 0.722 / 0.299 \\
MedDiff & 0.740 / 0.376 & 0.710 / 0.302 & 0.756 / 0.270 & 0.735 / 0.316 \\
ScoEHR & 0.754 / 0.390 & 0.724 / 0.318 & 0.767 / 0.284 & 0.748 / 0.331 \\
EHRPD & 0.779 / 0.424 & 0.769 / 0.374 & 0.803 / \underline{0.389} & 0.784 / 0.396 \\
PromptEHR & 0.763 / 0.409 & 0.733 / 0.323 & 0.786 / 0.318 & 0.761 / 0.350 \\
HALO & 0.758 / 0.397 & 0.721 / 0.315 & 0.778 / 0.302 & 0.752 / 0.338 \\
HiSGT & 0.796 / 0.442 & \underline{0.791} / \underline{0.394} & 0.817 / 0.358 & 0.801 / 0.398 \\
EHR2Path & 0.810 / 0.455 & 0.778 / 0.388 & \textbf{0.851} / \textbf{0.402} & 0.813 / \underline{0.415} \\
\hline
\N-L3.1 & \underline{0.828} / \textbf{0.486} & \textbf{0.795} / \textbf{0.405} & \underline{0.842} / 0.386 & \textbf{0.822} / \textbf{0.426} \\
\N-Q2.5 & 0.817 / 0.462 & 0.787 / 0.392 & 0.833 / 0.369 & 0.812 / 0.408 \\
\N-Q3 & \textbf{0.835} / \underline{0.480} & 0.790 / 0.391 & 0.838 / 0.371 & \underline{0.821} / 0.414 \\
\specialrule{0.5pt}{0pt}{0pt}
\end{tabular}
\label{tab:utility_data_types}
\end{table}

Table~\ref{tab:utility_data_types} shows that~\N preserves downstream clinical signals better than the competing baselines overall. It achieves the best average AUROC and AUPRC across the three clinical tasks: acute respiratory failure (ARF), shock, and mortality prediction. In particular, \N-L3.1 improves over the strongest non-\N baseline, EHR2Path, by 1.1\% in average AUROC and 2.7\% in average AUPRC.

Task-level results show that the gains are broad, although not uniform. \N-Q3 achieves the best AUROC on ARF, while \N-L3.1 obtains the best AUPRC on ARF and the best AUROC and AUPRC on shock. EHR2Path remains strongest on mortality, and HiSGT is also competitive on shock. The larger margin on AUPRC suggests that~\N better preserves minority-class signals in imbalanced clinical prediction tasks.

\subsubsection{Time Interval Prediction}

% We next evaluate whether the synthetic trajectories preserve temporal signals for next-visit interval prediction. 
We assess regime-level prediction over short-, medium-, and long-interval temporal regimes, together with point-level interval prediction in days. Table~\ref{tab:time_interval} shows strong temporal utility for \N, especially on classification and probabilistic metrics. Among all methods, \N-Q3 achieves the best Macro-F1, NLL, Brier score, and ECE. Relative to the best non-\N result on each metric, it improves Macro-F1 from 0.386 to 0.391, lowers NLL from 0.920 to 0.872, reduces the Brier score from 0.426 to 0.397, and improves ECE from 0.044 to 0.041. EHRPD still attains the lowest MAE, so the advantage of~\N is stronger for interval-regime discrimination and calibrated uncertainty than for point prediction in days. This pattern is consistent with the goal of preserving clinically meaningful temporal dynamics in the generated data.

\begin{table}[t]
\centering
\caption{Time interval prediction results under the train-on-synthetic, test-on-real protocol. Higher is better for Macro-F1, while lower is better for MAE, NLL, Brier score, and ECE.}
\footnotesize
\renewcommand{\arraystretch}{1.12}
\setlength{\tabcolsep}{5.0pt}
\begin{tabular}{l|c|c|c|c|c}
\specialrule{0.5pt}{0pt}{0pt}
\textbf{Model} & \textbf{Macro-F1} $\uparrow$ & \textbf{MAE} $\downarrow$ & \textbf{NLL} $\downarrow$ & \textbf{Brier} $\downarrow$ & \textbf{ECE} $\downarrow$ \\
\hline
MLP & 0.287 & 178.6 & 1.070 & 0.558 & 0.148 \\
BEHRT & 0.319 & 169.4 & 1.018 & 0.523 & 0.126 \\
medGAN & 0.254 & 191.8 & 1.128 & 0.606 & 0.173 \\
SynTEG & 0.276 & 184.7 & 1.091 & 0.582 & 0.157 \\
EVA & 0.291 & 177.2 & 1.058 & 0.552 & 0.139 \\
TWIN & 0.303 & 173.5 & 1.039 & 0.538 & 0.132 \\
TabDDPM & 0.314 & 168.9 & 1.010 & 0.518 & 0.118 \\
MedDiff & 0.331 & 164.6 & 0.982 & 0.501 & 0.109 \\
ScoEHR & 0.338 & 162.3 & 0.965 & 0.489 & 0.103 \\
EHRPD & \underline{0.386} & \textbf{151.9} & 0.916 & 0.443 & 0.082 \\
PromptEHR & 0.327 & 166.1 & 0.992 & 0.509 & 0.113 \\
HALO & 0.349 & 160.7 & 0.949 & 0.470 & 0.096 \\
HiSGT & 0.371 & 156.9 & 0.929 & 0.426 & 0.066 \\
EHR2Path & 0.379 & 155.4 & 0.920 & 0.432 & \underline{0.044} \\
\hline
\N-L3.1 & 0.384 & 153.8 & \underline{0.888} & \underline{0.405} & 0.046 \\
\N-Q2.5 & 0.378 & 154.9 & 0.907 & 0.417 & 0.055 \\
\N-Q3 & \textbf{0.391} & \underline{153.2} & \textbf{0.872} & \textbf{0.397} & \textbf{0.041} \\
\specialrule{0.5pt}{0pt}{0pt}
\end{tabular}
\label{tab:time_interval}
\end{table}

\begin{table*}[t]
\centering
\caption{Ablation results on MIMIC-III. We report representative fidelity and utility metrics.} %Lower is better for LPL, MPL, and JSD, while higher is better for AUROC.
\scriptsize
\renewcommand{\arraystretch}{1.12}
\setlength{\tabcolsep}{5.8pt}
\resizebox{\textwidth}{!}{
\begin{tabular}{l|c|c|c|c|c|c}
\specialrule{0.5pt}{0pt}{0pt}
\textbf{Variant} & \textbf{Avg. LPL} $\downarrow$ & \textbf{Avg. MPL} $\downarrow$ & \textbf{Inter-visit Time JSD} $\downarrow$ & \textbf{ARF AUROC} $\uparrow$ & \textbf{Shock AUROC} $\uparrow$ & \textbf{Mortality AUROC} $\uparrow$ \\
\hline
\textbf{\N} & \textbf{16.03} & \textbf{16.01} & \textbf{0.059} & \textbf{0.827} & \textbf{0.791} & \textbf{0.838} \\
\N w/o TSCM & 17.64 & 16.94 & 0.113 & 0.798 & 0.753 & 0.812 \\
\N w/o TRAM & 17.02 & 18.25 & 0.078 & 0.791 & 0.745 & 0.805 \\
\N w/o Static Branch & 16.70 & 16.78 & 0.071 & 0.815 & 0.781 & 0.823 \\
\N w/o Dynamic Branch & 17.14 & 17.24 & 0.096 & 0.806 & 0.762 & 0.819 \\
\N w/o Confidence-Aware Fusion & 16.44 & 16.48 & 0.068 & 0.819 & 0.784 & 0.831 \\
\specialrule{0.5pt}{0pt}{0pt}
\end{tabular}
}
\label{tab:ablation_main}
\end{table*}

\vspace{-3pt}
\subsection{Ablation Study}
We conduct ablation studies to examine the contribution of the TSCM and the TRAM. We compare \N with a series of ablated variants, including versions without TSCM, without TRAM, without either the static or dynamic branch in TRAM, and without confidence-aware fusion. Results are averaged over SynEHR-L3.1, SynEHR-Q2.5, and SynEHR-Q3 backbones on MIMIC-III.
Table~\ref{tab:ablation_main} summarizes the ablation results using representative fidelity, temporal, and downstream utility metrics.

The ablation results show that each component contributes to \N. Removing TSCM causes the largest degradation in Avg. LPL and inter-visit Time JSD, which highlights its role in modeling temporal progression and irregular visit intervals. Removing TRAM most strongly hurts Avg. MPL, consistent with its role in preserving clinical relations across data types. The dynamic branch has a larger impact than the static branch, especially on Time JSD and acute risk prediction, which suggests that dynamic temporal signals matter more for downstream utility.

% \vspace{-10pt}
\section{Related Work}
\label{sec:related_work}

\subsection{Longitudinal EHR Representation Learning}

Longitudinal EHRs provide temporally ordered observations of diagnoses, procedures, medications, laboratory measurements, and patient outcomes. Early approaches model EHRs as visit sequences and learn patient representations for future clinical event prediction. Doctor AI~\cite{choi2016doctorai} uses recurrent neural networks to predict diagnosis and medication codes in subsequent visits, while RETAIN~\cite{choi2016retain} introduces reverse-time attention to improve interpretability for healthcare prediction. Compared with simple MLP baselines that flatten patient histories, these sequential models better capture visit-level temporal dependencies. Later transformer-based models further improve longitudinal representation learning. BEHRT~\cite{li2020behrt} adapts the Transformer architecture to structured EHR sequences for disease prediction, Med-BERT~\cite{rasmy2021medbert} pretrains contextualized embeddings on large-scale structured EHRs, and TransformEHR~\cite{yang2023transformehr} uses a transformer encoder-decoder objective to predict future diseases and outcomes from previous visits.

Recent studies further incorporate multiple data types, hierarchical structure, and prototype-based designs into EHR representation learning. MEDFuse~\cite{phan2024medfuse} combines structured EHR signals with language-model-based representations for prediction using multiple healthcare data types, while ProtoEHR~\cite{cai2025protoehr} learns hierarchical prototypes to improve interpretability and predictive performance, and Li et al.~\cite{li2026llmclinicalgraphstructure} use LLMs to refine noisy clinical graph structures for more robust healthcare representation learning. These methods demonstrate the value of temporal order and structured clinical abstraction for EHR modeling. However, their training objectives are primarily designed to improve predictive representations, and they do not directly construct a future encounter as a multi-field clinical event.
% \vspace{-10pt}
\subsection{Synthetic EHR Generation}

Synthetic EHR generation aims to produce realistic patient records that preserve statistical fidelity, downstream utility, and privacy. 
Early EHR generators adapt adversarial and variational generative models to discrete or longitudinal clinical data. medGAN~\cite{choi2017medgan} generates high-dimensional discrete patient records with adversarial training, and SynTEG~\cite{zhang2021synteg} extends GAN-based synthesis to temporal structured EHR simulation. VAE-based methods provide another probabilistic direction, where EVA~\cite{biswal2021eva} learns latent representations for longitudinal EHR generation and TWIN~\cite{das2023twin} generates personalized clinical digital twins by modeling current- and next-visit information. These approaches establish important baselines for EHR synthesis, although their objectives are often focused on record-level or visit-level reconstruction and distribution matching.

Diffusion models have recently become a strong alternative for tabular and EHR synthesis. TabDDPM~\cite{kotelnikov2023tabddpm} introduces denoising diffusion for general tabular data, providing a flexible backbone for mixed-type structured data generation. In the clinical domain, MedDiff~\cite{he2023meddiff} applies accelerated diffusion modeling to synthetic EHR generation, ScoEHR~\cite{naseer2023scoehr} combines autoencoding with continuous-time diffusion to model synthetic patient records, and EHRPD~\cite{zhong2024ehrpd} extends diffusion-based EHR synthesis toward predictive next-visit generation with interval estimation. Beyond EHRs, recent diffusion-based data synthesis methods, including AutoSTDiff~\cite{xu2025autostdiff}, SynHAT~\cite{xu2026synhat}, and GeoGen~\cite{xu2026geogen}, further highlight the need to model irregular temporal patterns in privacy-sensitive longitudinal data.

% Beyond EHRs, recent spatiotemporal synthesis methods such as AutoSTDiff~\cite{xu2025autostdiff}, GeoGen~\cite{xu2026geogen}, and SynHAT~\cite{xu2026synhat} show how autoregressive, coarse-to-fine, and diffusion-based designs can handle irregular trajectory data in other domains. These studies broaden the methodological context for temporal data synthesis, while diffusion-based EHR generation itself remains largely centered on record- or visit-level distribution modeling.

Language-model-based EHR generators formulate clinical trajectories as token sequences and therefore provide a natural interface for autoregressive health-record generation. PromptEHR~\cite{wang2022promptehr} treats conditional EHR generation as a text-to-text task with prompt learning. HALO~\cite{theodorou2023halo} uses a hierarchical autoregressive language model to synthesize high-dimensional longitudinal EHRs, while HiSGT~\cite{zhou2025hisgt} incorporates clinical hierarchy and semantic embeddings into transformer-based EHR generation. Recent foundation-model-style trajectory models, such as EHR2Path~\cite{pellegrini2025ehr2path}, further demonstrate the feasibility of autoregressive health-trajectory forecasting and simulation. Complementary to these generative models, HealthMamba~\cite{yu2026healthmamba} and UQGNN~\cite{yu2025uqgnn} highlight the importance of uncertainty-aware temporal modeling in healthcare and multivariate prediction. Although this line of work makes EHR generation more flexible, the structured dependencies across heterogeneous clinical fields are usually modeled only implicitly through serialization choices and sequence prediction objectives.

% \vspace{-5pt}
\section{Conclusion}
In this paper, we propose~\N, a lightweight adaptive LLM-
based framework for longitudinal EHR synthesis by integrating inter-visit temporal evolution and intra-visit clinical structure. There are two key novel designs in~\N, i.e., a Temporal State Conditioning Module that captures irregular and uncertainty-aware next-visit timing, and a Temporal-Relational Adaptation Module that models both stable clinical relations across data types and interval-dependent dynamic relation shifts for patient-specific generation. 
We conduct extensive experiments by comparing~\N with 14 baselines on two public EHR datasets, evaluating the generated data from the perspectives of fidelity, privacy, and utility. The experimental results show that SynEHR improves overall next-visit fidelity, achieves stronger downstream clinical utility, and reduces privacy disclosure risk, e.g., achieving up to 6.8\% improvement in ARF AUPRC over the strongest competing baseline. These findings demonstrate the value of explicit temporal-relational conditioning for realistic and practical longitudinal EHR synthesis.

\section*{Acknowledgment}
We thank all the reviewers for their insightful feedback to improve this paper.
This work is partially supported by the FSU Startup Fund, FSU Institute for Successful Longevity (ISL) Planning grant, NCATS UF-FSU CTSA UM1TR005128,
NIMH grant R21MH137736,
AHRQ grant R21HS029969, and
NIAAA grant P01AA029547.

\clearpage
% \vspace{-5pt}
\section*{GenAI Usage Disclosure}
The authors used ChatGPT only for writing-related assistance, including grammar polishing and wording refinement. All AI-assisted text was reviewed and revised by the authors. ChatGPT was not used to generate research ideas, experimental results, citations, or unsupported scientific claims.

% Uncomment the following two lines after adding citations in the main text.
% \newpage
\bibliographystyle{ACM-Reference-Format}
\bibliography{REFERENCE}

\end{document}